Article

# Comparative Analysis of Kinect-Based and Oculus-Based Gaze Region Estimation Methods in a Driving Simulator [†]

David González-Ortega *, Francisco Javier Díaz-Pernas, Mario Martínez-Zarzuela and Míriam Antón-Rodríguez

Department of Signal Theory, Communications and Telematics Engineering, Telecommunications Engineering School, University of Valladolid, 47011 Valladolid, Spain; pacper@tel.uva.es (F.J.D.-P.); marmar@tel.uva.es (M.M.-Z.); mirant@tel.uva.es (M.A.-R.)
* Correspondence: davgon@tel.uva.es; Tel.: +34-983-423-000 (ext. 5552)
† This paper is a conference extension of González-Ortega, D.; González-Díaz, J.; Díaz-Pernas, F.J.; Martínez-Zarzuela, M.; Antón-Rodríguez, M. 3D Kinect-Based Gaze Region Estimation in a Driving Simulator. In Proceedings of the International Conference on Ubiquitous Computing and Ambient Intelligence, Philadelphia, PA, USA, 7–10 November 2017.

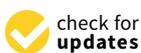

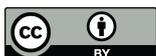

**Abstract:** Driver's gaze information can be crucial in driving research because of its relation to driver attention. Particularly, the inclusion of gaze data in driving simulators broadens the scope of research studies as they can relate drivers' gaze patterns to their features and performance. In this paper, we present two gaze region estimation modules integrated in a driving simulator. One uses the 3D Kinect device and another uses the virtual reality Oculus Rift device. The modules are able to detect the region, out of seven in which the driving scene was divided, where a driver is gazing at in every route processed frame. Four methods were implemented and compared for gaze estimation, which learn the relation between gaze displacement and head movement. Two are simpler and based on points that try to capture this relation and two are based on classifiers such as MLP and SVM. Experiments were carried out with 12 users that drove on the same scenario twice, each one with a different visualization display, first with a big screen and later with Oculus Rift. On the whole, Oculus Rift outperformed Kinect as the best hardware for gaze estimation. The Oculus-based gaze region estimation method with the highest performance achieved an accuracy of 97.94%. The information provided by the Oculus Rift module enriches the driving simulator data and makes it possible a multimodal driving performance analysis apart from the immersion and realism obtained with the virtual reality experience provided by Oculus.

## 1. Introduction

Event simulation is increasingly being used in many different areas. Simulators can speed up the process of acquisition of basic abilities and are configured as tools of great learning and research capacity in a wide range of areas such as surgery [1] or Internet of Things (IoT) [2]. Particularly, driving simulators make possible the learning and re-education of drivers through the inclusion of varied routes and traffic situations that can compromise safety [3]. Moreover, driving simulators store data that can be analyzed to study the different aspects that play a role on traffic safety and driving ability shortcomings that can lead to traffic dangerous situations.

With that aim, we developed a driving simulator with varied routes (urban and interurban) and traffic events to analyze the driving safety level. The simulator can show in real time and store in files for further processing, not only information about the vehicle (position, speed, rpm (revolutions per minute), gear, and fuel consumption) but also any committed traffic violations. We also developed an Android application that controls the Shimmer physiological sensors (electrocardiogram (ECG), electromyogram (EMG), and galvanic skin response (GSR)) through a Bluetooth connection [4]. The Android application is synchronized with the driving simulator, so that the physiological data can be used in driving research studies. We intended to extend the data extracted from the driver in our simulator to analyze his/her attention state.

An important aspect to measure driver allocation of attention is the gaze location and patterns [5,6]. Gaze dynamics is significant in many different driving scenarios. Hergeth et al. [7] stated that gaze monitoring provides a more direct measure of trust in autonomous driving than other behavioral approaches. Visual scanning behavior tends to narrow in intervals of high cognitive demand. Wang et al. [8] concluded in their research that horizontal gaze dispersion was the most sensitive information to measure changes in gaze concentration under cognitive demand. Many research studies have analyzed and proved the influence of age, physical fitness, and driving experience on gaze location and visual search strategies. Novice drivers adjust their visual search strategies to the environment situation less effectively than experienced drivers, gazing more often at the immediate surroundings, and relying less on peripheral vision for vehicle control. There is a relationship between the visual search strategies of novice drivers and their underdeveloped vehicle skills and less spare attentional capacity than experienced drivers [9,10]. Sun et al. [11] carried out an on-road driving study to assess visual-motor coordination in elderly drivers. Older drivers with better executive function skills performed more frequent eye fixations on the curves and inside vehicle features. Visual perception of the surroundings requires frequent gaze shifts partly caused by the fact that most of the area in the human field of view (FOV) falls upon peripheral vision [12]. Although small gaze shifts are mainly accompanied by the eyes and the head moves very little, for larger saccades the head contributes approximately 80% of the total change in gaze angle with the eyes contributing only approximately the remaining 20% [13]. Stiefelhagen et al. [14] studied head orientation and gaze direction in meetings and concluded that head has a contribution of about 70% in the gaze direction. Three reasons can be given to explain why we move our head in gaze shifts [15]. Firstly, head movements extend the range of gaze beyond the oculomotor range (about $\pm$ 55 deg) and also, by bringing the eyes back towards a central position, these head movements assist in apprehending future visual targets [16]. Secondly, compensatory head movements are made during body movements in space. Thirdly, head movements are also used to express emotions. Gaze shifts are usually composed of two stages [12]. In the first stage, the gaze is rapidly shifted to the targets using both the eyes and the head. After reaching the target, a gaze shift enters the second stage. In this stage, the head keeps on moving in the same direction adopted in the first stage, while the eyes move backwards with the same speed as the head. As a result, the gaze remains stabilized on the target. The two motor systems in charge of moving the eyes and the head receive the same command at almost the same time as natural mechanism that directs our gaze. This was concluded in an experiment studying eye-head coordination while driving [17]. Although eye and head motor systems are separate, there are clear interactions between them. During the eye movements carried out to gaze a new region, an analysis of the speed patterns identifies a first increase, then a decrease and later a new increase. All those changes are closely related to head movements [18,19]. Accordingly, some gaze estimation methods have adopted eye and head orientation, whereas others have utilized only head orientation [20]. Although by using eye and head orientation cues, detailed gaze direction can be estimated, eye orientation cannot always be measured in vehicular environments mainly caused by the occlusion of the eye region. There could be several reasons for this occlusion: sunlight reflections on glasses, blinks of the driver, partial occlusion of the pupil due to squinting, large head rotation, image blur, and vehicle vibration. Illumination changes, vehicle vibration, image blur, and poor video resolution can also affect the computation of the eye orientation. All of these factors can make eye orientation in a vehicle context inaccurate [20]. Moreover, the costs of high-resolution recording equipment and other computational requirements further enhance the difficulty of developing practical and deployable eye orientation solutions.

Coarse gaze direction can be obtained by using only head orientation since a driver's effective visual field is limited and, usually, a person moves the head to a comfortable position before orienting the eye [21]. The coarse gaze direction has been adopted in Forward Collision Warning (FCW) systems [22] or for gaze region estimation. Fridman et al. [23] presented a method to estimate driver gaze region using a 2D camera with information about only head position and not eye position. They showed that even small shifts in facial configuration are sufficiently distinct for a classifier to accurately disambiguate head pose into one out of six driver gaze regions. Pan et al. [24] presented a gaze tracking system with a web camera, and an ultrasonic sensor placed just above the camera and in its same plane. Gaze tracking is based on head movement in 3D space. Their system delivers outstanding performance. Depth cameras such as the Kinect device (Microsoft, Redmond, WA, USA) can provide robust head tracking for gaze estimation.

The Kinect sensor add-on for the Xbox 360 video game platform, which emerged at the end of 2010, interprets the 3D information of the scene obtained through infrared structured light that is read by a standard CMOS sensor. Its sampling frequency rate is 30 fps (frames per second). It is a low-cost alternative to 2D cameras that present problems difficult to address, such as occlusions between human body parts, range of movements, and drastic illumination and environmental changes. Since its commercial launch, many developers and researchers have used the Kinect device in their work, in different areas, e.g., head-pose and facial expression tracking, hand gesture recognition, human activity recognition, and healthcare applications [25,26]. Jafari and Ziou [27] presented a method for gaze estimation under normal head movement. While head position and orientation are acquired by Kinect, they obtained eye direction by a supplementary pan-tilt-zoom (PTZ) camera. Cazzato et al. [28] adopted the Kinect device for head pose estimation combining RGB and depth data in an unconstrained environment. The performance of their method was comparable to other approaches although they require significant constraints, invasive hardware, and supervised learning. Ghiass et al. [29] proposed a generic and fully automatic method to obtain the facial pose using the Kinect device. They also evaluated their method for 3D gaze estimation. It outperformed appearance-based gaze estimators, which require high resolution images, due to the high precision of the proposed head tracker. Virtual reality devices can also be used for head tracking and gaze direction estimation.

Virtual reality (VR) is common in video games and is increasingly present in other areas such as education and medicine [30]. VR can easily model a whole world in order to be used in teaching experiences immersing the user in its environment. Moreover, the new virtual reality techniques and devices achieve more immersive and realistic environments. Regarding teaching experiences, learning driving skills is undoubtedly a powerful application of virtual reality. Particularly, young drivers with limited experience can commit driving mistakes leading to fatal accidents. These young people are accustomed and receptive to new technologies such as virtual reality, through which they can be exposed to driving events in a safe and reliable environment. They can gain remarkable driving experience and raise awareness of the potential consequences of driving mistakes in order to change their dangerous behaviors. The inclusion of virtual reality devices (head mounted displays (HMD)) in driving simulators makes a difference. They have been adopted by most of the largest automotive manufacturers [31]. The great level of involvement and immersion provided by the latest HMDs can lead to a more natural interaction in a driving simulator, thus research on driver or even pedestrian behavior can benefit from them. Deb et al. [32] developed a pedestrian simulator with the Vive virtual reality device (HTC, New Taipei, Taiwan). Their simulator was successfully tested in a pedestrian crossing as it monitored the pedestrian behavior using head tracking and participants ranked it high in terms of usability and realism. Some VR devices can accomplish not only head tracking but also eye tracking. For instance, HTC Vive Pro Eye includes Tobii Eye Tracking Technology and HTC Vive Headset includes Pupil Labs Eye-Tracker add-on. However, Clay et al. [33] used the HTC Vive Headset and stated that it has to be calibrated and validated properly and that accuracy can vary depending on the eyes (e.g., bright eyes are better than dark eyes).

An outstanding virtual reality device is the Oculus Rift (Facebook Technologies, LLC, Menlo Park, CA, USA). Oculus Rift is an inexpensive 3D VR HMD released in 2012. It has 110° FOV in diagonal, a combined resolution of 1280 × 800, and an update frequency of 60 Hz. Oculus embeds micro-electrical-mechanical (MEMS) sensors and can track and monitor head movements so that the content display can be compensated in an immersive VR environment [34]. Although the Oculus Rift was originally designed for video games, researchers have explored the possibility of using it in areas such as rehabilitation [35] and driving simulators. Lhemedu-Steinke et al. [36] carried out an experiment comparing the evaluation of 84 participants driving in a simulator using Oculus Rift and a conventional display. The adoption of Oculus Rift enabled statistically significant better concentration, involvement, and enjoyment than a conventional display (a full HD smart TV). Ali et al. [37] also used the Oculus Rift in a gamified driving simulator to assess driving behaviors. Their experiments showed that 88% of the subjects considered that the use of the simulator was instructive and enabled them to improve driving behaviors. Actually, there was a 21% average increase in the correct actions and 17% average decrease in the wrong actions of the drivers after using the simulator with the Oculus Rift.

In the previously cited works using the Kinect [27–29], this device was used for gaze estimation although they have not only used head orientation for it. The use of additional devices and eye detection or tracking makes those approaches more complex and difficult to implement in vehicular environments. On the other hand, Oculus Rift and others HMDs have been used in driving research works [35–37] but, to the best of our knowledge, without including gaze region estimation, which can be accomplished using the head orientation obtained from MEMS sensors. The initial aim of our research was to extend the data acquired by our driving simulator with the acquisition of the regions that the driver is gazing at. For that purpose, we aimed to integrate different devices such as a depth camera (Kinect) and a VR HMD (Oculus Rift) in our driving simulator to estimate the region the driver is gazing at from the same data (head orientation) and the same features although this orientation is extracted from very different ways using both devices. Then, we aimed to compare the gaze region estimation performance using the two approaches and eventually to validate the Oculus Rift as the gaze region estimation device in a driving simulator and not only as the visualization device, which can significantly enrich the extracted data and therefore the analysis and scope of driving studies. We developed a Kinect-based gaze region estimation module and an Oculus-based gaze region estimation module and integrated them in our driving simulator. As the Oculus Rift has never been applied to gaze region estimation in driving research studies, the Oculus-based gaze region estimation module is an original contribution of the presented work. The modules are able to extract the region that the driver is gazing at each processed frame after an initial calibration stage. It is possible to select, before driving in a simulator scenario, the visualization device: a conventional display (big screen of dimensions 260 × 195 cm) or Oculus Rift. We have carried out experiments to compare the Kinect-based and Oculus-based gaze estimation performance. Such a comparison between a depth camera and a HMD for this task has never been proposed before and is another contribution of our work. We carried out an extensive comparison using the same four gaze estimation methods and the same features for each method with each device. The first two methods are bidimensional estimators using two head rotation angles (yaw and pitch). The difference between the two methods is the decision-maker applied as it is static in the first method and adapted in the second method. The third method uses an MLP classifier and the fourth a SVM classifier, both with the same features. The comparison has validated Oculus Rift for gaze region estimation in a driving simulator, which is another contribution of the presented work. The Oculus-based gaze region estimation can be applied to enrich the data extracted by VR applications from other areas such as rehabilitation and education. In [38], we presented preliminary results using only Kinect and not Oculus Rift for gaze region estimation and with only one of the four methods analyzed in this work.

The rest of the paper is organized as follows: Section 2 describes the developed 3D Kinect-based and Oculus-based gaze region estimation modules that we integrated in a driving simulator. Experimental results are detailed in Section 3. Finally, Section 4 draws the main conclusions about the presented work.

## 2. Materials and Methods

### 2.1. Kinect-Based and Oculus-Based Gaze Region Estimation

We aimed to integrate a gaze region estimation module in a driving simulator that we had already developed. The driving simulator includes different scenarios and traffic situations interesting to analyze the safety level of the drivers in the experiments. The driver has to react to all the situations adequately to ensure safety. The simulator stores a lot of data from the driver vehicle on each route such as position, speed, rpm, gear, and fuel consumption. The simulator was developed with the Unity game engine (Unity Technologies, San Francisco, CA, USA) and the 3D modeling programs Blender (Blender Foundation, Amsterdam, The Netherlands) and 3ds Max (Autodesk, San Rafael, CA, USA). It can be executed in an off-the-shelf computer and a low-cost input device such as the G27 (Logitech, Lausanne, Switzerland). The simulator allows to analyze the driving style of different people taking into account their reactions to the traffic situations they have to face and their outcomes. It also allows the awareness-raising and the learning of safe driving techniques.

The joint analysis of the gaze patterns, driving styles, and safety level of drivers can identify the visual search strategies that lead to the safest driving. Moreover, the training on these visual search strategies included in the simulator can improve visual search skills in drivers of varied age and experience as stated by [39].

Driver's gaze can be characterized through different representations such as directional vector [40], 3D points [41], dynamic objects of interest [42], and static regions of interest [6]. Many approaches utilize static regions of interest [43–46] as this representation allows one to extract high semantic information such as fixations in the different regions and saccades from one region to another. From this information, driver attention state, level of drowsiness, engagement in secondary activities, and prediction of intended maneuvers can be computed. The normal driving is characterized by gazing straight ahead most of the time with temporal fixations on the mirrors that changes when engaged in secondary tasks. A driver may perform the secondary task with one long glance away from the forward driving direction or may perform the secondary tasks via multiple short glances towards the region involved in the secondary task (e.g., a GPS navigation display) [47]. On the other hand, driving gazing away from the forward direction becomes increasingly dangerous as long as time goes by and the driver does not gaze straight ahead [40]. Moreover, the static gaze regions allow to extract high semantic information and then to compute driver attention state and other data previously mentioned in this section.

We have adopted two different approaches to estimate the gaze region in the simulator, each one with a different visualization device. First, we use a big screen of dimensions 260 × 195 cm as the visualization device and the 3D Kinect device to estimate the gaze region. Then, we use the Oculus Rift device as both the visualization and the gaze estimation device.

For the Kinect, we have used the Kinect for Windows SDK and the Face Tracking SDK for Kinect for Windows in the driving simulator Unity project to integrate a module in our simulator to achieve Kinect-based head tracking and gaze region estimation from it.

When a user is in front of the Kinect device, the gaze estimation module achieves head tracking through the extraction of a large number of facial points. If these points are joined forming triangles, the 3D facial mesh is built. The module tracks robustly the head in frontal and non-frontal positions. From those facial points, three rotation angles are obtained: yaw, pitch, and roll. Yaw is the head rotation around the vertical axis, pitch is the head rotation around the side-to-side axis, and roll is the head rotation around the front-to-back axis.

We estimate the gaze region through the head position. From experimental results, we discarded the roll angle because it does not provide relevant information regarding the driver's head position when he or she is gazing at the different regions in which we have divided the simulator scene. The yaw and pitch angles are the basic values used for gaze estimation in the different adopted methods, which are explained in the next section.

With Kinect, the simulator scene is projected onto a big screen of dimensions 260 × 195 cm. The distance between the screen and the driver is 250 cm. The increase of the yaw angle to move the head from the place when the user is gazing at the central position of the screen, in order to gaze at the right border of the screen, is 27.47 deg supposing that the eyes remain in the same position that they have while gazing at the central position of the screen. That value can be obtained taking into account that the tangent of this angle equals the division between half the screen width and the distance between the screen and the user (see Figure 1). Consequently, the hardware horizontal FOV is 54.94 deg. The software horizontal FOV is 100 deg. We selected the screen distance of 250 cm after feedback from users in previous experiments. They found driving more comfortably at this distance to the screen although hardware and software horizontal FOV do not match. The hardware FOV determines the range of head movement necessary to gaze at the different driving scene regions. We have divided the screen into seven representative regions: (1) straight ahead, (2) speedometer, (3) rear-view mirror, (4) left mirror and left door glass, (5) steering wheel with another speedometer, (6) rpm meter and gear indicator, and (7) right part of the dashboard, as shown in Figure 2.

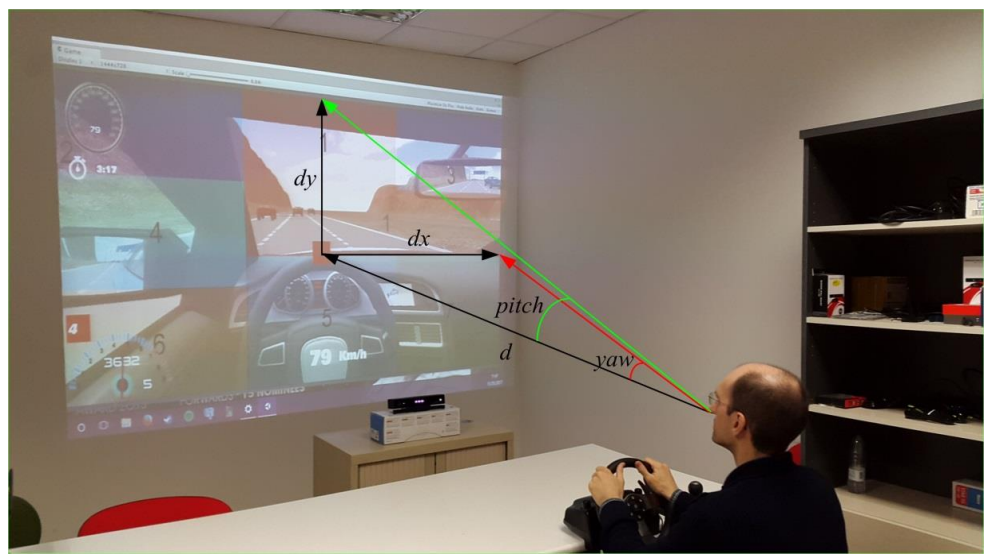

**Figure 1.** Transformation from degrees of yaw and pitch to cm in the visualization screen.

We will use numbers 1 to 7 to refer to these regions throughout the paper. It was very important to take a good decision about the number of scene regions as it is necessary to distinguish among all the regions that the driver is going to gaze for a different purpose but taking into account that the larger the number of regions is, the worse the gaze region estimation performance is. Many research studies used similar gaze regions both in simulation and on-road studies with small differences to adapt to the particular environment and FOV [43–46]. For instance, Thorslund et al. [43] selected seven quite similar gaze regions in a driving simulation research study, including windshield, left, rear-view mirror, speedometer, secondary task display, and other (region with objects not included in the other regions). Vora et al. [44] considered six quite similar regions, including forward, left, center stack, rear-view mirror, and speedometer, in an on-road experiment to analyze the performance of a gaze region estimation approach using convolutional neural network.

With Oculus as a visualization device, the scene is rendered twice, one with the left camera and another with the right camera, in such a way that each eye watches the corresponding scene, producing the three-dimensional scene with the fusion of both scenes inherent in human stereoscopic vision. We have integrated an Oculus-based gaze region estimation module in our driving simulator that uses the information of the yaw and pitch head rotation angles. The scene is divided in the same seven regions previously mentioned when the visualization device is the big screen. Figure 3 shows a user with Oculus Rift in the driving simulator. Figure 4 show images that a user, who is driving different vehicles and using Oculus Rift, is watching with one eye. Changes in the FOV as a function of head movements can be observed. Images in Figure 4 differ from the 3D images that the driver is really watching with Oculus Rift through the fusion of the images watched in each eye, achieving an immersive and realistic experience.

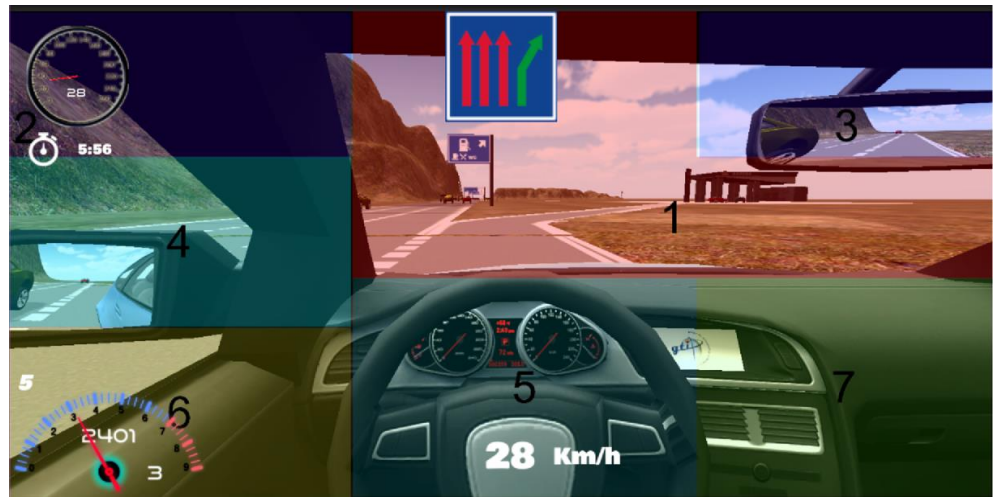

**Figure 2.** Simulator scene with the seven regions considered for the gaze estimation analysis.

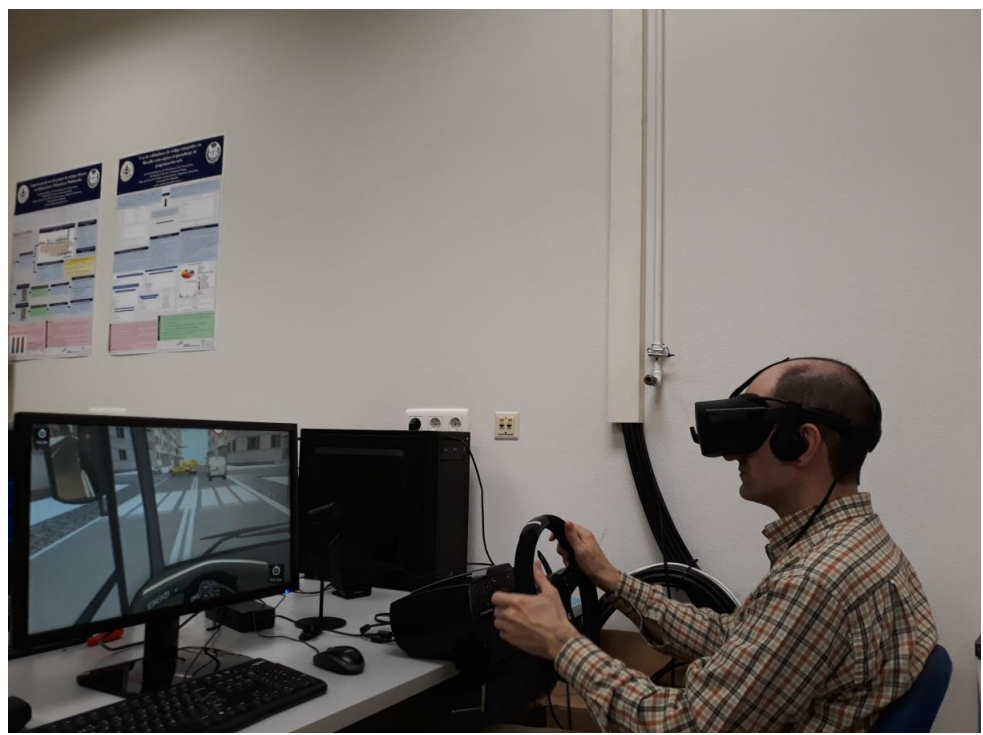

**Figure 3.** User driving in the simulator with Oculus Rift.

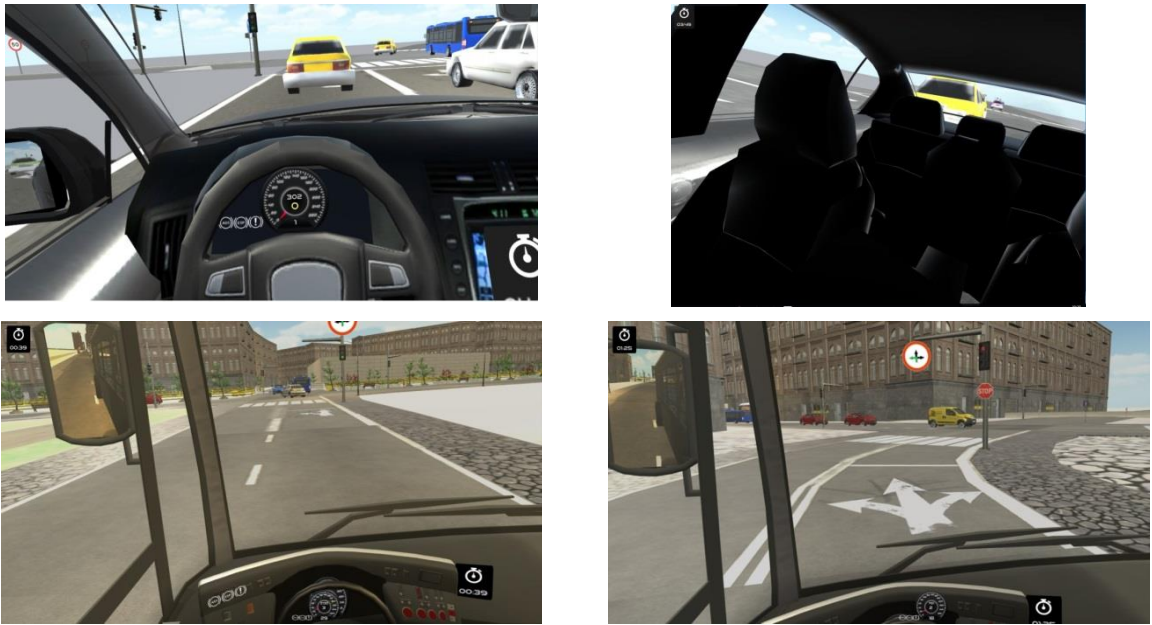

**Figure 4.** Images watched using Oculus Rift as the visualization display in the simulator.

*2.2. Methods Implemented for Gaze Region Estimation*

We have implemented four different methods for gaze estimation and have applied them using the Kinect device and the Oculus Rift.

2.2.1. Method 1: Bidimensional Estimation with Static Decision-Maker

This method uses the yaw and pitch angles. Roll was discarded as it is a movement not naturally made in a driving scenario as mentioned in [22]. On the other hand, we are assuming that the user will have a head movement following a gaze displacement. Anyway, we quantify this feature individually in the calibration.

Both yaw and pitch initially range from 0 to 360 deg increasing rightwards and upwards. We have subtracted 360 deg to the value of yaw and pitch if they range from 180 to 360 deg. This way, they range from −180 to 180 deg. The pitch angle has a positive value if the head is looking up the ceiling and has a negative value if the head is looking down the floor. The yaw angle has a positive value if the head is turned towards the right shoulder and it has a negative value if the head is turned towards the left shoulder. Although absolute values near 90 deg are difficult to track, the range of values for a person driving in the simulator are much smaller and then precisely obtained by the Kinect.

Once the axes representing *yaw* (x axis) and *pitch* (y axis) are transformed, the displacement in the x axis (*dx*) and in the y axis (*dy*) of the head rotation can be calculated as shown in Equations (1) and (2):

$$dx = tan(yaw) \cdot d \quad (1)$$

$$dy = tan(pitch) \cdot d \quad (2)$$

where $tan(yaw)$ and $tan(pitch)$ are the tangents of the *yaw* and *pitch* angles, respectively. *yaw* and *pitch* are the rotation angles with respect to the central position (0,0) in the axes. *d* is the known distance between the user and the screen. In Figure 1 these values are graphically represented.

To compensate for the individual differences between head rotations and gaze displacements, a calibration stage to configure the decision maker is included in the method. In this stage, the user has to gaze at five different points (one central point in blue color and four red points on the borders) in the screen as shown in Figure 5. The values of *dx* and $dy$, which are obtained while the users are gazing at these five points, are considered the training set as they are used to configure the structure of the decision maker.

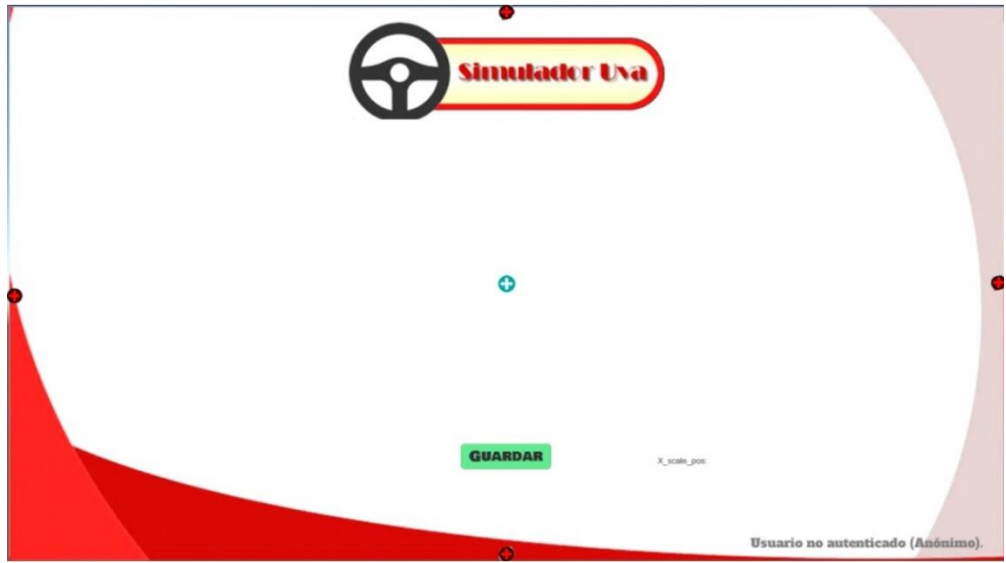

**Figure 5.** Calibration points in method 1.

The values of $dx$ and $dy$ obtained while the user is gazing at the central point (0,0) of the screen are used to modify the displacements in all the points, such that:

$$dx = dx - dx_{(0,0)} \qquad (3)$$

$$dy = dy - dy_{(0,0)} \qquad (4)$$

The values of $dx$ while gazing at pt1 (red point on the right in Figure 5) and pt2 (red point on the left in Figure 5) are used to obtain the scaling factors $s_{x+}$ and $s_{x-}$, respectively:

$$s_{x+} = \frac{dx_{pt1}}{\frac{w}{2}} \qquad (5)$$

$$s_{x-} = \frac{dx_{pt2}}{\frac{w}{2}} \qquad (6)$$

where $w$ is the width of the screen.

Similarly, the values of $dy$ while gazing at pt3 (red point at the top in Figure 5) and pt4 (red point at the bottom in Figure 5) are used to obtain the scaling factors $s_{y+}$ and $s_{y-}$, respectively:

$$s_{y+} = \frac{dy_{pt3}}{\frac{h}{2}} \qquad (7)$$

$$s_{y-} = \frac{dy_{pt4}}{\frac{h}{2}} \qquad (8)$$

where $h$ is the height of the screen.

After the calibration, the value of $dx$ and $dy$ will be calculated with the Equations (9) and (10), respectively, depending on the positive or negative value of $dx$ and $dy$:

$$dx = \left\{ \begin{array}{l} dx \cdot s_{x+}, dx > 0 \\ dx \cdot s_{x-}, dx < 0 \end{array} \right\} \qquad (9)$$

$$dy = \left\{ \begin{array}{l} dy \cdot s_{y+}, dy > 0 \\ dy \cdot s_{y-}, dy < 0 \end{array} \right\} \qquad (10)$$

Depending on $dx$ and $dy$, we use the decision maker considering the sizes of the regions shown in Figure 6. For instance, region 3 is estimated if $dx$ is larger than $0.2 \cdot w$ and $dy$ is larger than $0.1 \cdot h$.

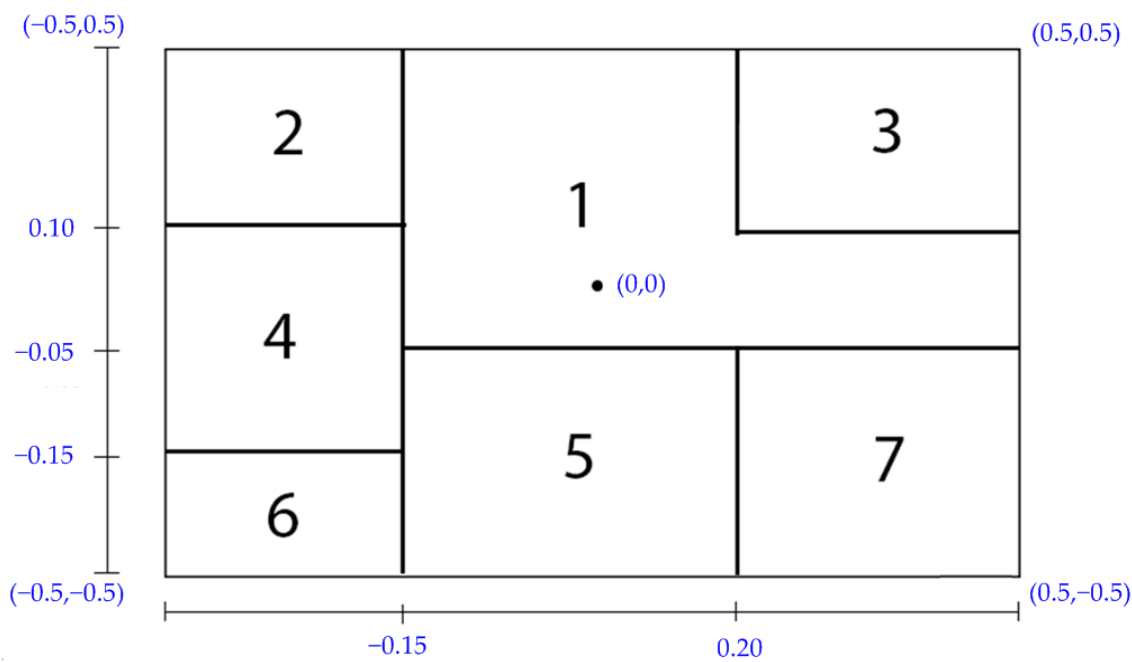

**Figure 6.** Structure of the decision maker as a function of the different regions.

2.2.2. Method 2: Bidimensional Estimation with Adapted Decision Maker

This method is a modification of the first method, in which not only the relation between the head movement and the gaze displacement is considered, but also the different way the users are gazing at the different regions. This problem was considered following a statement present in [48], which is that the way of gazing at different regions is not directly related to the head movement but relatively related to the turning degrees necessary for a person to see an object within their FOV.

In this method, the calibration stage to configure the decision maker consists in gazing at a series of points distributed along the border of the different regions in which we have divided the scene, as shown in Figure 7. From the 23 points, displacements in the x axis are only considered for the points from 1 to 12 and displacements in the y axis are only considered for the points from 13 to 23. The user has to gaze at each point in a natural way, after having gazed at the central point, to establish a relation between the head movement and the gaze displacement in each point. $dx$ and $dy$, which are obtained while the users are gazing at these 23 points, are considered the training set as they are used to configure the structure of the decision maker.

Once calibration with all the points is fulfilled, the decision maker is built comparing $dx$ and $dy$ of the head movement in each frame with their values in the calibrated points. For instance, region 3 is estimated if $dx$ is larger than $dx$ at point 11 and $dy$ is larger than $dy$ at point 22.

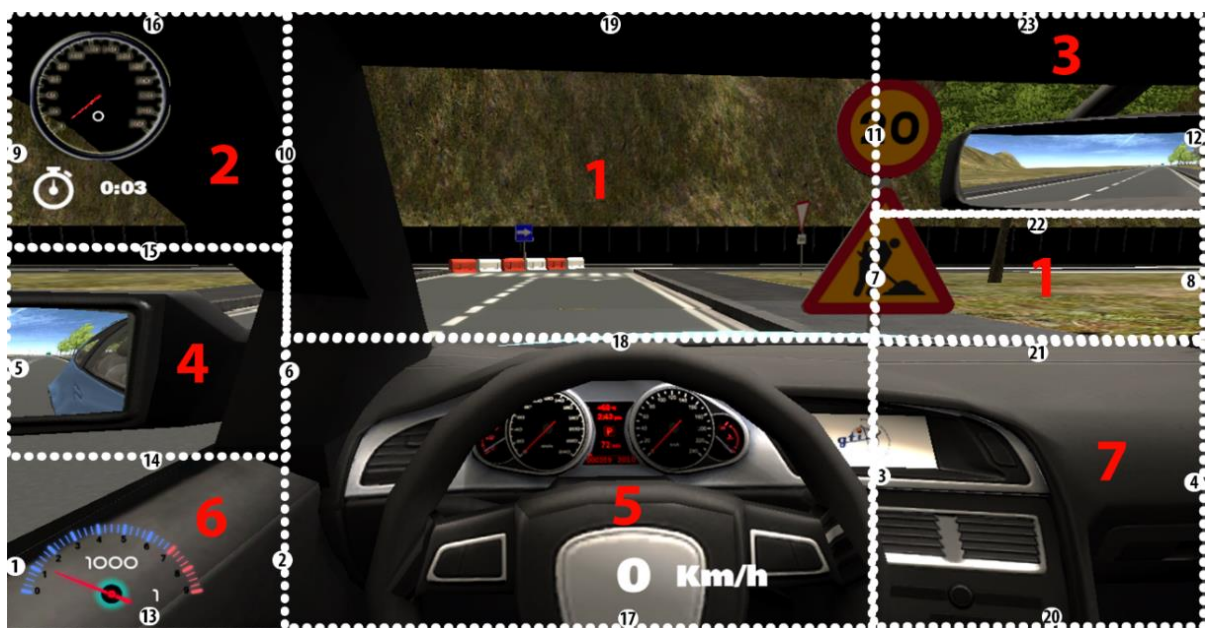

**Figure 7.** Calibration points along the border of the seven regions (numbered 1 to 7) used in method 2.

2.2.3. Method 3: MLP Classifier

In this method, we adopted a multi-layer perceptron (MLP), which has been successfully applied to a great variety of problems ranging from time series forecasting to image classification. MLP is a feedforward artificial neural network (ANN) whose units or neurons are organized in layers: one input layer, one or more hidden layers, and one output layer. The values of the features of an input pattern are transferred from the input layer to the hidden layers and then to the output layer in a feedforward fashion. The layers are fully connected and the value of each neuron depends on the value of the neurons of the previous layer, the weights of the connections with that layer, and an activation function [49]. The most important parameters in an MLP are the weights and biases. These are obtained in the training stage and they determine that the MLP achieves the desired output for the input patterns. The backpropagation algorithm is used in the training stage to adjust the weights by sending the difference between the calculated output and the expected output, called the cost function, backward through the network [50].

We adopted the C++ implementation of MLP provided by the Machine Learning Library (MLL), which is included in Open Source Computer Vision Library (OpenCV). We have used an architecture comprising of seven binary MLPs in the method. All the MLPs have a hidden layer comprised of 14 neurons. Each MLP is in charge of detecting that the user is gazing at one, out of the seven regions of the scene. Once a pattern is processed for the seven MLPs, the detected gaze region will be given by the MLP with the largest output value.

As input parameters to each MLP, we selected the parameters used in [13], namely:

(1) Yaw.
(2) Pitch.
(3) Displacement in the x axis of the central point of the rectangle given by the face detection in each frame. This displacement is relative to the value in the x axis of the central point of the rectangle given by the face detection when the user is asked to gaze at the central point of the scene.
(4) Displacement in the y axis of the central point of the rectangle given by the face detection in each frame. This displacement is relative to the value in the y axis of the central point of the rectangle given by the face detection when the user is asked to gaze at the central point of the scene.
(5) Area of the rectangle given by the face detection in each frame.

With these five features, we applied a feature selection algorithm to analyze the discriminative power of them. We adopted the $J_5(\xi)$ criterion [51]. Following this criterion, the correlation between the interclass and the intraclass distances is calculated to obtain the separability between classes. Equation (11) shows that $J_5(\xi)$ is the trace ($tr$) of the matrix result of the product of the inverse of $S_w$, which is the average of the covariance matrices inside each class, and the $S_b$, which is the covariance matrix for the averages of each class:

$$J_5(\xi) = tr\left\{ S_W^{-1} \cdot S_b \right\} \tag{11}$$

The classification of the features using this criterion from the most to the least discriminative is: 1, 2, 3, 4, and 5 (preserving the number in which they were previously mentioned) with weights 36.44, 68.35, 71.36, 72.85, and 74.23, respectively. The first weight represents the discriminative power of the first feature, the second weight represents the discriminative power of the first and second features, the third weight represents the discriminative power of the first, second, and third features, and so on.

To generate the patterns for the MLP training in the calibration stage, each user has to gaze at each region 10 s moving their eyes over the region to generate 150 patterns. Each pattern has the previously mentioned five features: yaw, pitch, displacement in the x axis of the central point of the rectangle given by the face detection in each frame, displacement in the y axis of the central point of the rectangle given by the face detection in each frame, and area of the rectangle given by the face detection in each frame. After gazing at each region, the user has to gaze at the central point of the scene for two reasons: firstly, to consider the influence of the initial position of the head on its movement to gaze at another region, and secondly, to calculate the mean value of the $dx$ and $dy$ of the head position in the central point of the scene used to compute the 3rd and 4th feature of each pattern. These 150 patterns per region and per user are used to train the MLPs.

The configuration of the seven MLPs of the adopted architecture are similar. The value of their training parameters is:

- Maximum value of iterations: 1000.
- Convergence parameter epsilon: 0.001.
- Training method: Backpropagation.
- Strength of each gradient in the weight computation: 0.1.
- Difference between weights of two consecutive iterations: 0.1.

The selected activation function is the symmetric sigmoid.

2.2.4. Method 4: SVM Classifier

The classifier used in this method is the support vector machine (SVM). It is a pattern classification technique applied in many different machine learning areas, such as generalized predictive control and image classification [52]. Due to its flexibility and generalization ability, SVM achieves balance predictive performance. SVM aims to project separable samples onto a high dimensional space with the adoption of kernel functions and where a separating hyperplane that maximizes the distance from the closest data points can be found. With compact datasets, the selection of a linear kernel as opposed to a nonlinear one both decreases the risk of overfitting and improves the classification performance.

Similarly to the implementation of MLP, we adopted the C++ implementation of SVM provided by the MLL, which is included in OpenCV. The steps for the creation and configuration of the classifier are similar to the steps in method 3, although, instead of using seven classifiers in the system, it is comprised of just one SVM classifier, which has as output the region, out of the seven, where the user is gazing at in each frame. Thus, to generate the patterns for the SVM training in the calibration stage, each user has to gaze at each region 10 s moving their eyes over the region to generate 150 patterns. Each pattern has the same five features used for method 3: yaw, pitch, displacement in the x axis of the central point of the rectangle given by the face detection in each frame, displacement in the y axis of the central point of the rectangle given by the face detection in each frame, and area of the rectangle given by the face detection in each frame. These 150 patterns per region and per user are used to train the SVM. The selected training parameters are:

- Maximum value of iterations: 100,000.
- Convergence parameter epsilon: 0.001.

The selected classifier parameters are:

- Parameter controlling the outlier separation: 1.
- Kernel type: Linear.

## 3. Experimental Results

We evaluated the Kinect-based and the Oculus-based gaze region estimation using the four methods explained in the previous section. Twelve subjects took parts in the experiments. The mean age of the subjects ranged from 20 to 55, with a mean of 27.83. The mean years of driving experience was 8.75 years, ranging from less than a year to 38 years. Seven of the subjects had experience as video game players while the remaining five subjects did not have such experience.

Many significant research studies about driver gaze region estimation carried out their experiments with 12 or fewer subjects, such as [6,22,39,44,45,53]. We performed a sample size estimation with paired *t*-tests [54] using the SPSS statistical package (version 24, IBM, Armonk, NY, USA) to prove that a sample size of 12 subjects would be sufficient to detect differences between the Kinect-based gaze region estimation and the Oculus-based gaze region estimation for each subject and using the same methods. Paired t-tests were carried out with the differences in gaze region estimation using Kinect and Oculus for each of the four developed methods. The t-tests showed that a sample size of 10 subjects would have been enough to detect differences between the Kinect-based gaze region estimation and the Oculus-based gaze region estimation with a significant level ($\alpha$) f 0.05 and power ($1 - \beta$) of 0.95, although we extended the sample size to 12 subjects in our experiments.

Each driver made the same route in our driving simulator in the research laboratory twice: first using a screen as the visualization display and another time using Oculus Rift as the visualization display. We decided that users would use Oculus Rift the second time they drove on the route. With VR devices such as Oculus Rift, simulator sickness has been reported that could force users to withdraw from the experiments [32]. Consequently, users were asked to drive the second time using the Oculus Rift. In any case, no user suffered from simulator sickness in the experiments and thus all the users were able to finish the experiments. Each of the seven regions of interest of the scene was colored differently to the surrounding regions and identified with a printed number to make the participants as easy as possible to say the region they are gazing at, which is necessary to test the implemented methods. The color of the regions and the size and position of the numbers were carefully selected to ease the identification of the regions without compromising the driving performance. Figure 6 shows the numbered regions. Figure 2 shows the regions numbered and colored as the drivers gazed at them in the experiments. Regions 1, 2, and 4 were colored reddish, violetish, and greenish, respectively. Regions 6 and 7 were colored yellowish. Each user had to complete a calibration stage for each method and device, which was explained in the previous section together with the explanation of each implemented method. In the calibration stage, the training set utilized to configure the method was created. After that, the user had to drive on the route and had to say out loud the number of the region that was gazing at every time a red square appeared in the middle of the screen. That red square appeared every 4 s on the screen during the simulation route. The drivers were asked to gaze at the different scene regions in a natural way and according to the requirements to drive safety on the simulator route. To react properly to the road, traffic signs, road traffic, and other events present on the route properly, they had to gaze at the different representative regions in which we divided the scene. When the drivers were required to say the region they were gazing at, they may make a gaze displacement to focus on the digit to make sure that they said the region they were gazing at before. Besides, the researcher who was annotating the region said by the driver and also the region given by our approach, considered the possible delay and gaze displacement between the time the driver was asked to say the region he/she is gazing at and when he actually said it as this probably implicated that the driver had to gaze at the number of the region. In this case, the researcher annotated the region given by our approach previously to the gaze displacement. From the feedback given by the drivers after the experiments, we also have to mention that, in the calibration stage, the drivers were familiarized with the position and number of the different regions and somehow learned that information. Therefore, they drove naturally on the route and said the scene region easily and without interfering on the driving. On the bottom left part of the screen below the left mirror, the region detected by the module is written so that we wrote it down each pair of regions for a processed frame: the actual region that the driver is gazing at and the region detected by the module. Figure 8 shows simulator scenes with the seven considered regions and the number of the region detected by the Kinect-based gaze region estimation module printed in white below the left mirror.

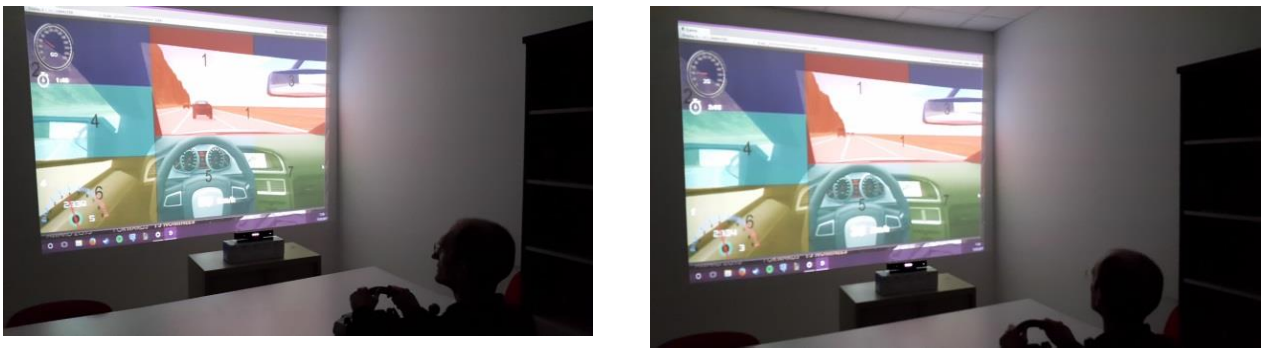

**Figure 8.** Simulator scenes using Kinect-based gaze region estimation.

Approximately 500 pairs of regions were collected for each participant and method in the Kinect-based gaze estimation and another 500 pairs in the Oculus-based gaze estimation. Each pair is composed by the actual region the participant was gazing at and the region predicted by the method for the same image. The testing set is composed by the patterns formed from the processing of the images corresponding to the moments when each participant had to say out loud the number of the region that was gazing at every 4 s. These patterns have 2 features ($dx$ and $dy$) for methods 1 and 2 and 5 features (yaw, pitch, displacement in the x axis of the central point of the rectangle given by the face detection in each frame, displacement in the y axis of the central point of the rectangle given by the face detection in each frame, and area of the rectangle given by the face detection in each frame) for methods 3 and 4. The performance of each device and method was computed comparing each pair of regions associated to the same image. There was a correct prediction if both regions are the same and there was a wrong prediction if the regions were different."

While in the training set all the regions were equally considered to learn the relation between gaze displacements and head orientation, in the testing set we used images where drivers were gazing at the regions in a natural way (corresponding to their driving on the route) and thus the actual regions the drivers were gazing at are not equally distributed. The region 1 (straight ahead) was by far the most gazed at region and regions 3 (rear-view), 4 (left mirror and left door glass), and 5 (steering wheel with another speedometer) were more gazed than regions 2 (speedometer), 6 (rpm meter and gear indicator), and 7 (right part of the dashboard). The regions more present in the testing set were the regions more important to drive properly on the simulator route. To obtain the overall accuracy of each method, we considered all the images equally in the testing set, instead of averaging the accuracy for all the regions. Therefore, the gazed regions more present in the images, as they are more important for the driving, had more influence on the overall accuracy.

We aimed to compute the performance of the methods this way as the usefulness of the gaze region estimation module in the driving simulator would then be higher in further experiments to analyze the gaze patterns of drivers as a function of their features and performance. For this reason, we discarded to use in the testing set images where drivers were gazing at all the regions with the same frequency as this approach would have not been representative of real and natural driving.

Before analyzing the Kinect-based and Oculus-based performance of the different methods, we provide some statistics about yaw and pitch of the users while driving in the experiments. Table 1 shows statistics about yaw and pitch obtained from Kinect and using the previously mentioned screen as the visualization display. Table 2 shows the same statistics obtained from Oculus Rift. With both Kinect and Oculus Rift, the range of values of yaw are larger than the range of values of pitch, which is expected as left and right turns are more needed than upward and downward turns. The range of values of yaw using Kinect is smaller than using Oculus. The difference between the largest and the lowest yaw considering all the users is 49.34 deg with Kinect and 84.40 with Oculus, being in both cases the largest and lowest values quite similar in absolute value. Contrastively, the difference between the largest and lowest pitch considering all the users is 24.42 deg with Kinect and 32.94 with Oculus Rift. The standard deviation of yaw is much lower with Kinect (8.76 deg) than with Oculus (19.27 deg), which can lead to a more discriminative power of yaw with Oculus Rift to distinguish among the different gaze regions. The standard deviation of pitch is also larger with Oculus (6.80 deg) than with Kinect (4.29 deg) although their values are relatively close. There are great differences between the largest value for one user and the lowest value for one user of the computed statistics obtained from Kinect and Oculus Rift. This highlights the great individual dissimilarities in the head movements to gaze at the different regions while driving and thus the importance of including a calibration stage in gaze estimation methods.

**Table 1.** Statistics about yaw and pitch obtained from Kinect.

|  | Mean Value Considering all the 12 Users | Largest Value for One User | Lowest Value for One User |
|---|---|---|---|
| Mean yaw (deg) | 0.33 | 3.54 | −3.53 |
| Standard deviation of yaw (deg) | 8.76 | 13.10 | 2.84 |
| Largest yaw (deg) | 23.50 | 35.40 | 10.77 |
| Lowest yaw (deg) | −25.84 | −11.17 | −50.30 |
| Mean variation of yaw per second (deg/s) | 2.64 | 4.39 | 0.77 |
| Mean pitch (deg) | 10.73 | 16.54 | −2.72 |
| Standard deviation of pitch (deg) | 4.29 | 7.40 | 2.68 |
| Largest pitch (deg) | 20.67 | 29.70 | 11.61 |
| Lowest pitch (deg) | −3.75 | 4.76 | −13.38 |
| Mean variation of pitch per second (deg/s) | 1.16 | 2.51 | 0.61 |

**Table 2.** Statistics about yaw and pitch obtained from Oculus Rift.

|  | Mean Value Considering all the 12 Users | Largest Value for One User | Lowest Value for One User |
|---|---|---|---|
| Mean yaw (deg) | −2.51 | 3.53 | −7.80 |
| Standard deviation of yaw (deg) | 19.27 | 32.60 | 12.96 |
| Largest yaw (deg) | 40.57 | 50.97 | 25.17 |

Table 2. *Cont*.

|  | Mean Value Considering all the 12 Users | Largest Value for One User | Lowest Value for One User |
|---|---|---|---|
| Lowest yaw (deg) | −43.83 | −29.75 | −62.43 |
| Mean variation of yaw per second (deg/s) | 2.83 | 5.16 | 1.30 |
| Mean pitch (deg) | −3.26 | 3.00 | −9.04 |
| Standard deviation of pitch (deg) | 6.80 | 11.98 | 4.05 |
| Largest pitch (deg) | 8.87 | 16.92 | 2.22 |
| Lowest pitch (deg) | −24.07 | −14.13 | −34.71 |
| Mean variation of pitch per second (deg/s) | 0.80 | 1.24 | 0.37 |

In the next subsections, first, the results for the four methods using the Kinect are presented. Then, the same methods using the Oculus Rift are presented.

*3.1. Kinect-Based Gaze Region Estimation*

The confusion matrices for the 7-region gaze estimation problem using each of the four developed methods and for all the users are presented in Table 3. The row dimension is indexed by the actual regions the drivers were gazing at and the column dimension is indexed by the detected regions given by the gaze estimation.

Using the bidimensional method with static decision-maker (method 1), the global accuracy including all the regions is 88.79%. All the regions except region 3 have an accuracy higher than 84%. The regions with higher accuracy are region 1, which has the largest area and is on the center of the image, and regions 2, 6, and 7, which are relatively large and placed on the corners of the scene. From these four regions, region 7 has the highest accuracy. The region with the lowest accuracy is region 3, which is the smallest region and farthest from the center of the scene. The individual results of three users yield interesting results: the user that mostly accompanied the gaze displacement with the head movement, the user with average head movement, and the user that accompanied the gaze displacement with the smallest head movement. The results of the user with the biggest head movement were coherent with the overall results. Although his/her overall accuracy is considerably larger (94.92%), the smallest accuracy percentage for region 3 (74%) is quite large, being the smallest and most cornered region. Accuracy in region 4 is considerably large (97%) compared to the accuracy in region 4 for all the users (84%), due to the pronounced head movement when he/she was looking gazing at this region. The results of the user with the average generic head movement are similar to the results considering all the users, although the regions 2 and 5 have particularly good results (95%). The user with the smallest head movement has an overall accuracy of 84.84% with particular bad results in region 3 (44%). It can be observed that users with small head movements have large error rates in region 3. The second region with the largest errors is region 4 as it is quite small and has several surrounding areas.

Using the bidimensional estimation with adapted decision maker (method 2), the global accuracy is 92.85%, 4.06% larger than with method 1. Moreover, the errors are distributed more regularly among all the regions. The regions with the smallest error are regions 1 and 5, which are on the center of the scene. The error rate of the user with the largest head movement is below 4%. The only region with accuracy smaller than 96% for this user is region 3, with 84%. The global accuracy of the user with average head movement is even better (96.61%) than the accuracy of the user with the biggest head movement. Method 2 approximates the results of all the users achieving a large overall accuracy. The calibration stage in this method is better adapted to each user as the different way a user moves his/her head to gaze at each region is considered. Region 3 is not the one with the smallest accuracy as regions 4 and 6 have smaller accuracy. The accuracy of the user with the smallest head movement has increased 4.45% from method 1. Regions 3 and 4 are by far the ones with the smallest accuracy (59% and 69%) as they are small and the user made a small head movement to gaze at them.

Table 3. Confusion matrices using Kinect for each of the four developed methods.

| **Accuracy: 88.79%** **Error Rate: 11.21%** | | \multicolumn{7}{c}{**Kinect-Based Gaze Region Estimation (Method 1)**} |
|---|---|---|---|---|---|---|---|---|
| | | **1** | **2** | **3** | **4** | **5** | **6** | **7** |
| | 1 | 94% | 0% | 0% | 0% | 5% | 0% | 0% |
| | 2 | 1% | 93% | 0% | 6% | 0% | 0% | 0% |
| | 3 | 36% | 0% | 64% | 0% | 0% | 0% | 0% |
| Actual region | 4 | 4% | 8% | 0% | 84% | 0% | 3% | 0% |
| | 5 | 12% | 0% | 0% | 0% | 87% | 0% | 0% |
| | 6 | 0% | 0% | 0% | 8% | 2% | 90% | 0% |
| | 7 | 5% | 0% | 0% | 0% | 0% | 0% | 95% |
| **Accuracy: 92.85%** **Error Rate: 7.15%** | | \multicolumn{7}{c}{**Kinect-Based Gaze Region Estimation (Method 2)**} |
| | | **1** | **2** | **3** | **4** | **5** | **6** | **7** |
| | 1 | 98% | 0% | 0% | 0% | 1% | 0% | 0% |
| | 2 | 8% | 92% | 0% | 0% | 0% | 0% | 0% |
| | 3 | 21% | 0% | 79% | 0% | 0% | 0% | 0% |
| Actual region | 4 | 15% | 5% | 0% | 79% | 0% | 1% | 0% |
| | 5 | 4% | 0% | 0% | 0% | 96% | 0% | 0% |
| | 6 | 3% | 0% | 0% | 6% | 3% | 88% | 0% |
| | 7 | 9% | 0% | 1% | 0% | 11% | 0% | 79% |
| **Accuracy: 64.79%** **Error Rate: 35.21%** | | \multicolumn{7}{c}{**Kinect-Based Gaze Region Estimation (Method 3)**} |
| | | **1** | **2** | **3** | **4** | **5** | **6** | **7** |
| | 1 | 60% | 6% | 5% | 14% | 11% | 3% | 1% |
| | 2 | 9% | 70% | 2% | 17% | 1% | 1% | 0% |
| | 3 | 16% | 2% | 70% | 3% | 2% | 0% | 8% |
| Actual region | 4 | 12% | 11% | 1% | 65% | 2% | 8% | 1% |
| | 5 | 14% | 1% | 2% | 7% | 72% | 4% | 1% |
| | 6 | 11% | 2% | 0% | 16% | 5% | 66% | 0% |
| | 7 | 6% | 0% | 7% | 4% | 8% | 1% | 73% |
| **Accuracy: 64.73%** **Error Rate: 35.27%** | | \multicolumn{7}{c}{**Kinect-Based Gaze Region Estimation (Method 4)**} |
| | | **1** | **2** | **3** | **4** | **5** | **6** | **7** |
| | 1 | 64% | 4% | 5% | 11% | 14% | 1% | 1% |
| | 2 | 12% | 68% | 1% | 17% | 1% | 0% | 0% |
| | 3 | 17% | 1% | 68% | 2% | 3% | 0% | 9% |
| Actual region | 4 | 14% | 12% | 1% | 58% | 4% | 11% | 1% |
| | 5 | 19% | 1% | 2% | 9% | 68% | 1% | 1% |
| | 6 | 15% | 2% | 0% | 21% | 5% | 56% | 0% |
| | 7 | 9% | 0% | 4% | 1% | 6% | 0% | 79% |

With the MLP classifier (method 3), the global accuracy is 64.79%, much worse than the accuracy obtained in the first two methods. In general, the detection error in all the regions is quite similar, so the classifier is balanced. Differently from the results in methods 1 and 2, there are errors between not adjacent regions. These errors should be caused by outliers in the training stage. Regarding the results from the user with the largest head movement, the user with average head movement, and the user with the smallest head movement, they follow the expected trend although there are not satisfactory results in any of them. We analyzed the performance of the MLP-based classifier using only the yaw and pitch features, similarly to methods 1 and 2. The results are shown in Table 4. The overall accuracy decreases 5.73% (59.05%). The detection errors are sensitively worse for regions 4 and 6. It can be said that the other 3 features used in the classifier improve the performance and removing them is not worth it as the increase in the processing time is not important.

**Table 4.** Confusion matrix using method 3 (only two features: yaw and pitch) and Kinect.

| Accuracy: 59.05% Error Rate: 40.95% | | Kinect-Based Gaze Region Estimation | | | | | | |
|---|---|---|---|---|---|---|---|---|
| | | 1 | 2 | 3 | 4 | 5 | 6 | 7 |
| Actual region | 1 | 69% | 4% | 5% | 4% | 15% | 2% | 1% |
| | 2 | 28% | 53% | 2% | 10% | 2% | 3% | 1% |
| | 3 | 21% | 3% | 60% | 3% | 8% | 4% | 3% |
| | 4 | 24% | 14% | 3% | 41% | 9% | 7% | 1% |
| | 5 | 24% | 2% | 6% | 3% | 57% | 5% | 2% |
| | 6 | 20% | 3% | 7% | 24% | 11% | 28% | 7% |
| | 7 | 11% | 3% | 13% | 2% | 17% | 3% | 52% |

With the SVM classifier, the global accuracy is 64.73%, very similar to method 3. The regions with the largest detection error are 4 and 6, when these regions in method 3 are 1 and 4. The accuracy for the user with the largest head movement is 87.60%. The region with the largest detection error (27%) is region 2, which is a region that is not frequently gazed at. The accuracy for the user with average head movement is 53.01%, worse than the overall accuracy considering all the users. The user with the smallest head movement has a very large error rate (50.61%), even larger than the accuracy. We tested the SVM system using only the features yaw and pitch (Table 5). The overall accuracy is similar (only 0.47% worse) to the obtained using 5 features. The 5 features are useful to detect smaller regions of the scene although the central region is more frequently confused with the lateral regions in return.

**Table 5.** Confusion matrix using method 4 (only two features: yaw and pitch) and Kinect.

| Accuracy: 64.25% Error Rate: 35.75% | | Kinect-Based Gaze Region Estimation | | | | | | |
|---|---|---|---|---|---|---|---|---|
| | | 1 | 2 | 3 | 4 | 5 | 6 | 7 |
| Actual region | 1 | 74% | 3% | 4% | 2% | 14% | 1% | 2% |
| | 2 | 18% | 63% | 2% | 10% | 4% | 2% | 1% |
| | 3 | 16% | 2% | 61% | 1% | 9% | 1% | 9% |
| | 4 | 26% | 13% | 4% | 43% | 9% | 4% | 1% |
| | 5 | 28% | 3% | 6% | 2% | 58% | 1% | 2% |
| | 6 | 16% | 5% | 2% | 24% | 14% | 37% | 2% |
| | 7 | 13% | 1% | 10% | 1% | 12% | 1% | 60% |

*3.2. Overall Comparison of the Kinect-Based Methods*

Table 6 shows the comparison of the estimation accuracy of each gaze region for the four Kinect-based methods and all the users. While methods 1 and 2 have similar relative behavior regarding the accuracy of each region, they differ from method 3 and 4. Region 1 has large accuracy for methods 1 and 2 (94% and 98%, respectively), while this region in one of the regions with lower accuracy for methods 3 and 4. Region 3 has low accuracy for methods 1 and 2, while it has relatively large accuracy for methods 3 and 4. Table 7 shows a comparative summary of the four methods. Methods 1 and 2 are notably better than methods 3 and 4. The latter methods have quite similar results. Processing time in methods 1 and 2 are insignificant although methods 3 and 4 are fast enough for our system. Calibration for method 1 is very fast and easy, being increasingly difficult for method 2 and for methods 3 and 4. The latter methods (3 and 4) need to generate all the necessary training patterns. Moreover, the calibration in methods 1 and 2 can be controlled so calibration mistakes can be corrected.

**Table 6.** Comparative results of the estimation accuracy of each gaze region for the four Kinect-based methods and all the users.

|  | Method 1 | Method 2 | Method 3 | Method 4 |
|---|---|---|---|---|
| Region 1 | 94% | 98% | 60% | 64% |
| Region 2 | 93% | 92% | 70% | 68% |
| Region 3 | 64% | 79% | 70% | 68% |
| Region 4 | 84% | 79% | 65% | 58% |
| Region 5 | 87% | 96% | 72% | 68% |
| Region 6 | 90% | 88% | 66% | 56% |
| Region 7 | 95% | 79% | 73% | 79% |
| Overall accuracy | 88.79% | 92.85% | 64.79% | 64.73% |

**Table 7.** Comparative summary of the four Kinect-based methods.

|  | Method 1 | Method 2 | Method 3 | Method 4 |
|---|---|---|---|---|
| Overall error rate | 11.21% | 7.15% | 35.21% | 35.27% |
| Error rate for the user with the largest head movement | 5.07% | 3.63% | 17.21% | 12.39% |
| Error rate for the user with average head movement | 9.89% | 3.38% | 29.91% | 46.98% |
| Error rate for the user with the smallest head movement | 15.15% | 10.70% | 53.04% | 50.61% |
| Error rate using 2 features (yaw and pitch) | 11.21% | 7.15% | 40.94% | 35.74% |
| Calibration ease | High | High | Medium | Medium |
| Calibration time | 30′ | 1′30″ | 3′00″ | 3′00″ |
| Calibration supervision and correction | Yes | Yes | No | No |

In summary, method 2 is the most powerful method as it has an easy calibration and the best results. After that, method 1 is the second best. The third is the SVM system as it can work with relatively good results with only two features. The worst method is the MLP-based system although it has a considerable ability to adapt to any type of user.

*3.3. Oculus-Based Gaze Region Estimation*

The confusion matrices for the 7-region gaze estimation problem using each of the four developed methods and for all the users are presented in Table 8. Using the bidimensional method with static decision-maker (method 1), the global accuracy is 96.62%. All the regions have an accuracy larger than 93% except region 4 with 81%. The accuracy is 7.83% higher than in method 1 using the Kinect.

Table 8. Confusion matrices using Oculus for each of the four developed methods.

| Accuracy: 96.62% Error Rate: 3.38% | | Oculus-Based Gaze Region Estimation (Method 1) | | | | | | |
|---|---|---|---|---|---|---|---|---|
| | | 1 | 2 | 3 | 4 | 5 | 6 | 7 |
| Actual region | 1 | 100% | 0% | 0% | 0% | 0% | 0% | 0% |
| | 2 | 1% | 95% | 0% | 4% | 0% | 0% | 0% |
| | 3 | 7% | 0% | 93% | 0% | 0% | 0% | 0% |
| | 4 | 3% | 4% | 0% | 81% | 0% | 11% | 0% |
| | 5 | 3% | 0% | 0% | 0% | 96% | 0% | 0% |
| | 6 | 0% | 0% | 0% | 2% | 1% | 97% | 0% |
| | 7 | 1% | 0% | 0% | 0% | 0% | 0% | 99% |
| **Accuracy: 97.94% Error Rate: 2.06%** | | **Oculus-Based Gaze Region Estimation (Method 2)** | | | | | | |
| | | 1 | 2 | 3 | 4 | 5 | 6 | 7 |
| Actual region | 1 | 100% | 0% | 0% | 0% | 0% | 0% | 0% |
| | 2 | 2% | 96% | 0% | 2% | 0% | 0% | 0% |
| | 3 | 5% | 0% | 95% | 0% | 0% | 0% | 0% |
| | 4 | 2% | 3% | 0% | 89% | 0% | 6% | 0% |
| | 5 | 2% | 0% | 0% | 0% | 98% | 0% | 0% |
| | 6 | 0% | 0% | 0% | 2% | 0% | 98% | 0% |
| | 7 | 1% | 0% | 0% | 0% | 0% | 0% | 99% |
| **Accuracy: 59.95% Error Rate: 40.05%** | | **Oculus-Based Gaze Region Estimation (Method 3)** | | | | | | |
| | | 1 | 2 | 3 | 4 | 5 | 6 | 7 |
| Actual region | 1 | 55% | 1% | 19% | 17% | 5% | 0% | 3% |
| | 2 | 5% | 46% | 15% | 27% | 3% | 2% | 1% |
| | 3 | 8% | 1% | 74% | 7% | 3% | 1% | 7% |
| | 4 | 6% | 8% | 6% | 64% | 4% | 11% | 1% |
| | 5 | 15% | 0% | 1% | 10% | 60% | 4% | 9% |
| | 6 | 2% | 2% | 2% | 18% | 7% | 67% | 1% |
| | 7 | 4% | 0% | 1% | 3% | 4% | 0% | 88% |
| **Accuracy: 87.48% Error Rate: 12.52%** | | **Oculus-Based Gaze Region Estimation (Method 4)** | | | | | | |
| | | 1 | 2 | 3 | 4 | 5 | 6 | 7 |
| Actual region | 1 | 97% | 1% | 1% | 1% | 1% | 0% | 0% |
| | 2 | 6% | 84% | 0% | 7% | 0% | 2% | 0% |
| | 3 | 13% | 0% | 83% | 1% | 1% | 0% | 2% |
| | 4 | 10% | 24% | 1% | 62% | 0% | 2% | 0% |
| | 5 | 23% | 1% | 0% | 1% | 74% | 0% | 0% |
| | 6 | 2% | 1% | 0% | 24% | 0% | 71% | 0% |
| | 7 | 5% | 0% | 9% | 1% | 2% | 0% | 83% |

Using the bidimensional estimation with adapted decision maker (method 2), the global accuracy is 97.94%, better than in method 1. Moreover, region 4 has the smallest accuracy (89%) but with an increase of 8% with respect to method 1. The accuracy is 5.09% larger than in method 2 using the Kinect.

Using the MLP classifier (method 3), the global accuracy is 59.95%, much worse than in methods 1 and 3. Moreover, the accuracy is 4.84% worse than the result of the same method using Kinect unlike the comparison of methods 1 and 2 using the Kinect and Oculus. The largest region (1) has worse accuracy (55%) than smaller regions such as 3, 4, and 6.

Using the SVM classifier (method 4), the global accuracy is 87.48%, much better than method 3 but worse than methods 1 and 2. Differently to what happened with method 3 comparing the use of the Kinect and Oculus, the overall accuracy of method 4 using the Oculus is 22.75% better than using the Kinect.

*3.4. Overall Comparison of the Oculus-Based Methods*

Table 9 shows the comparison of the estimation accuracy of each gaze region for the four Oculus-based methods and all the users. While methods 1 and 2 have similar relative behavior regarding the accuracy of each region, they differ from method 3 and 4. The most marked divergence is between methods 1 and 2, on the one hand, and method 3, on the other hand. Region 1 has 100% accuracy for methods 1 and 2, while it is the region with the second lowest accuracy for method 3. While region 3 has the second lowest accuracy for methods 1 and 2, it has the second largest accuracy for method 3.

**Table 9.** Comparative results of the estimation accuracy of each gaze region for the four Oculus-based methods and all the users.

|                  | **Method 1** | **Method 2** | **Method 3** | **Method 4** |
| ---------------- | ------------ | ------------ | ------------ | ------------ |
| Region 1         | 100%         | 100%         | 55%          | 97%          |
| Region 2         | 95%          | 96%          | 46%          | 84%          |
| Region 3         | 93%          | 95%          | 74%          | 83%          |
| Region 4         | 81%          | 89%          | 64%          | 62%          |
| Region 5         | 96%          | 98%          | 60%          | 74%          |
| Region 6         | 97%          | 98%          | 67%          | 71%          |
| Region 7         | 99%          | 99%          | 88%          | 83%          |
| Overall accuracy | 96.62%       | 97.94%       | 59.95%       | 87.48%       |

*3.5. Overall Comparison of Results Using Oculus and Kinect*

The Kinect-based gaze region estimation module processes 30 fps, whereas the Oculus-based gaze region estimation module processes 60 fps. The dissimilar frame rates are given by the features of the Kinect and Oculus Rift devices. This difference did not affect the comparative analysis of both modules as the performance tests are based on the results given by the developed methods (the predicted region) and the ground-truth (the region the driver is actually gazing at) every 4 s regardless of the frame rate of each device.

FOV and mostly the distance between regions are larger with the Oculus than with the screen as visualization device. Consequently, head movements to gaze at the different regions are larger, which had some influence on the better performance of the methods with the Oculus than with the Kinect with all the users regardless the degree of head movement in each gaze displacement and the use of peripheral vision. The results are also caused by the precision of the head orientation given by the MEMS sensors embedded in the Oculus Rift and by the way the users move their eyes and head in a VR environment such as the Oculus Rift in both the calibration stage and on route, more similarly than using the screen as the visualization device.

The system that uses seven MLPs is the only method that has worse performance with the Oculus. A change in the configuration of the MLP-based system may have changed the performance although we have maintained the same configuration used with the Kinect to make fair comparisons. Training patterns and training process are fundamental in the performance of the MLP and SVM methods. The performance of methods 1 and 2 are quite sound with a difference in accuracy for region 4 (small region surrounded by other regions), sensitively better in method 2. Calibration takes more time and is more complex in methods 3 and 4 than in methods 1 and 2, which also have the best performance.

To assess the Oculus-based driving simulation, the users were asked to rate two aspects: the level of immersion and realism and the level of smooth natural head tracking provided by the Oculus Rift. The users had to rank both aspects quantitatively ranging from 0 (the worst rating) to 10 (the best rating). The average score for the two aspects was very satisfactory: 8 for the level of immersion and realism and 9 for the level of smooth natural head tracking.

*3.6. Statistical Analysis of the Order of the Experiments*

As it was already stated, the order of the devices used in the experiments was the same for all the subjects: first with the Kinect and using a screen as the visualization display and later using the Oculus Rift as the visualization display. Although we did it to avoid possible simulation sickness using Oculus, the order of the experiments may have affected the results. Using first the Oculus and then Kinect in the experiments for a subset of the subjects may have led to different gaze region estimation accuracies than using first the Kinect and then the Oculus. To test the possible influence of the order of the experiments, we have repeated the experiments with a subset of seven of the available subjects that took part in the already presented results. In the repeated experiments, the subjects drove in the same scenario twice, first using the Oculus Rift to estimate the gaze region and then using the Kinect. We have fulfilled a one-way analysis of variance (ANOVA) using the SPSS statistical package to test the influence of the order of the experiments in the subtraction between the Kinect-based and the Oculus-based gaze region estimation accuracies. Let G1 be the groups of experiments using first the Kinect and then the Oculus and let G2 be the groups of experiments using first the Kinect and then the Oculus. Before the ANOVA test, Levene's test for equality of variances was performed. The obtained significance level p-value was 0.439 ($p > 0.05$). Therefore, there is no difference between G1 and G2 in the variance of the subtraction between the Kinect-based gaze estimation and the Oculus-based gaze region estimation accuracies and the two groups are homogeneous with respect to the computed variable. In the ANOVA analysis, the significant level p-value was 0.744 ($p > 0.05$) so there are no significant differences between the value of the subtraction between the Kinect-based gaze estimation and the Oculus-based gaze region estimation accuracies between the two groups. This way, it was proved that the order in the use of the gaze estimation device did not affect the obtained gaze region estimation accuracies.

## 4. Conclusions

We have developed Kinect-based and Oculus-based gaze region estimation modules integrated in a driving simulator so that gaze region can be stored in the simulator and used in the analysis together with other data from the vehicle and the driver. We carried out a novel and original comparative analysis between both modules. Four different methods were implemented and compared to estimate gaze region through head movement in both modules. The Oculus Rift achieved an overall better performance than the Kinect due to different reasons such as the larger distance between regions with the Oculus, the precision of the head orientation from MEMS sensors, and the different features of head movements the drivers made to gaze at the different scene regions. The Oculus-based gaze region estimation method with the highest performance uses a bidimensional estimation with adapted decision maker. This method achieved an accuracy of 97.94% and outperformed MLP and SVM. The Oculus Rift, which makes driving experience more immersive and realistic, can be used without any additional hardware to obtain the region the driver is gazing at every frame in future research studies to analyze the gaze pattern of drivers as a function of their features and correct them reeducating the drivers if these patterns are not suitable for safe and efficient driving. The Oculus-based gaze region estimation provided by the method with the highest performance can be applied to other areas where virtual reality is used such as education or clinical therapies. The extraction of the region the user is gazing at while using a virtual reality application enriches the computed information and enables further analysis relating his/her performance in a VR application and his/her visual attention to the different stimulus present in the VR scenario.

Our approach does not obtain eye tracking to estimate gaze position but head tracking. We considered that eye tracking requires restrictions that cannot always be met in vehicular environments. Moreover, some virtual reality devices such as the Oculus Rift do not support eye tracking and the necessary introduction of other device providing eye tracking between the user and the Oculus Rift would greatly complicate its applicability and adoption. Therefore, the use of head movement to obtain gaze data is particularly interesting in driving and virtual reality environments. To achieve a precise gaze region estimation from head tracking, a calibration stage is included in our modules to adapt to the individual relation between head rotations and gaze displacements. This calibration process is fast and has to be carried out only once even though a driver uses the simulator repeatedly.

After the validation of the Oculus-based gaze region estimation module, we will use it in our diving simulator for experiments with many different scenarios and drivers to analyze the relation between the profile of the drivers, their gaze patterns, reactions to significant traffic events, and level of safe and efficient driving. We plan to adapt the Oculus-based gaze region estimation approach to other VR applications we developed in the area of physical and cognitive rehabilitation to enrich the extracted information from the use of the applications with the patients' gaze patterns.

**Author Contributions:** Conceptualization, D.G.-O. and F.J.D.-P.; Methodology, D.G.-O. and F.J.D.-P.; Software, D.G.-O., M.M.-Z. and M.A.-R.; Experimental results, D.G.-O., M.M.-Z. and M.A.-R., Supervision, D.G.-O.; Writing—original draft, D.G.-O., F.J.D.-P, M.M.-Z. and M.A.-R. All authors have read and agreed to the published version of the manuscript.

**Funding:** This work was funded by the National Department of Traffic (DGT, Dirección General de Tráfico) of the Ministry of the Interior (Spain) under research project SPIP2015-01801.

**Institutional Review Board Statement:** The study was conducted according to the guidelines of the Declaration of Helsinki. Ethical review and approval were waived as we complied with the regulations regarding participants' safety and privacy and data treatment was anonymous and confidential.

**Informed Consent Statement:** Informed consent was obtained from all subjects involved in the study.

**Data Availability Statement:** Data sharing not applicable.

**Conflicts of Interest:** The authors declare no conflict of interest.